\documentclass[letterpaper, 10 pt, conference]{ieeeconf}
\usepackage{tikz}
\usepackage{algorithmic}
\usepackage{algorithm}
\usepackage{graphicx}
\usepackage{amssymb}
\usepackage{amsmath}
\usepackage{amsthm}
\usepackage[compact]{titlesec}
\usepackage{cite}
\usepackage{caption}
\usepackage{multirow}
\usepackage{adjustbox}
\usepackage{caption}
\usepackage{subcaption}

\IEEEpeerreviewmaketitle

\DeclareMathOperator*{\argmax}{argmax}
\DeclareMathOperator*{\argmin}{argmin}

\IEEEoverridecommandlockouts                              % This command is only needed if you want to use the \thanks command

\renewcommand{\baselinestretch}{0.925} 

 \newtheorem{theorem}{\textbf{Theorem}}[section]

 \newtheorem{proposition}[theorem]{Proposition}
 \newtheorem{corollary}[theorem]{Corollary}

 \newtheorem{remark}{Remark}

\usepackage{textcomp}
\def\BibTeX{{\rm B\kern-.05em{\sc i\kern-.025em b}\kern-.08em
    T\kern-.1667em\lower.7ex\hbox{E}\kern-.125emX}}
\begin{document}
\title{Model Extraction Attacks Against Reinforcement Learning Based Controllers
}

\author{Momina Sajid, Yanning Shen, and Yasser Shoukry
    % <-this % stops a space
\thanks{The authors are with the Department of Electrical Engineering and Computer Science, University of California, Irvine, CA 92697, USA (e-mail: \{msajid,yannings,yshoukry\}@uci.edu).}
\thanks{This work was partially sponsored by the NSF award \#CNS-2002405 and C3.AI Digital Transformation Institute.}
}
\maketitle
\begin{abstract}
We introduce the problem of model-extraction attacks in cyber-physical systems in which an attacker attempts to estimate (or extract) the feedback controller of the system. Extracting (or estimating) the controller provides an unmatched edge to attackers since it allows them to predict the future control actions of the system and plan their attack accordingly. Hence, it is important to understand the ability of the attackers to perform such an attack. In this paper, we focus on the setting when a Deep Neural Network (DNN) controller is trained using Reinforcement Learning (RL) algorithms and is used to control a stochastic system. We play the role of the attacker that aims to estimate such an unknown DNN controller, and we propose a two-phase algorithm. In the first phase, also called the offline phase, the attacker uses side-channel information about the RL-reward function and the system dynamics to identify a set of candidate estimates of the unknown DNN. In the second phase, also called the online phase, the attacker observes the behavior of the unknown DNN and uses these observations to shortlist the set of final policy estimates. We provide theoretical analysis of the error between the unknown DNN and the estimated one. We also provide numerical results showing the effectiveness of the proposed algorithm.

\end{abstract}

%%% Outline of the final report
\section{Introduction}
Cyber-Physical Systems (CPS) control the vast majority of today's critical infrastructure and these CPSs are vulnerable to attacks that range from denial-of-service~\cite{6623750}, reply attacks~\cite{mo2009secure}, man-in-the-middle attacks~\cite{9482668}, and false data injection attacks~\cite{shoukry2013non}. Given the safety-criticality nature of CPSs, it is unsurprising to see the current interest to develop control-theoretic techniques to protect feedback controllers in these systems against such attacks~\cite{shoukry2015event,dan2010stealth,shoukry2021secure}.

Motivated by these vulnerabilities, in this paper, we introduce the model extraction attacks on CPSs. In machine learning literature, model extraction attacks refer to a set of algorithms that aims to extract (or estimate) the parameters of complex Deep Neural Network (DNN) models by observing input-output data collected from these DNNs.
These attacks have been widely studied in the supervised learning setup~\cite{shokri2017membership,tramer2016stealing}. In the CPS context, DNNs are widely used as feedback controllers thanks to the recent success in using Deep Reinforcement Learning (DRL) to train such complex feedback controllers. In the DRL setting, gradient-descent-based algorithms aim to train a DNN that maximizes a given reward function. In this spirit, we study model extraction attacks on feedback control systems which can play a significant role in understanding the vulnerabilities of CPSs. First, while in some cases the control algorithms themselves are sophisticated and proprietary, the more common case is that the algorithm parameters constitute the intellectual property as they are based on careful modeling, analysis, training, and validation processes at often considerable expense of time and resources. Hence, extracting such controllers provides significant loss in terms of learning intellectual properties. Second, extracted feedback controllers can be reverse engineered to identify weaknesses in CPSs, predict their future behavior, and launch complex attacks.

In this paper, we play the role of an attacker whose goal is to estimate (or extract) an unknown DNN controller using input-output observations collected from the DNN along with side-channel information about the reward function used to train the DNN. 

Several technical challenges arise in this setup. First, the attacker cannot control the quality and quantity of the input-output data collected from the unknown DNN. In scenarios where the amount of input-output observation is relatively small, or these observations do not cover the entire space of input-outputs, it is hard to construct function estimators that provide small bounded error guarantees away from the provided input-output samples. Second, the reward function provides limited knowledge about the unknown DNN. This stems from the fact that DRL algorithms aim to solve non-convex problems (due to the nonlinearity of the DNNs). Hence, the exact parameters of the DNN depend on a set of unknown parameters like the exact optimization algorithm used, the initial DNN parameters, the learning rates, the stopping criteria, and other hyperparameters. 

To that end, we propose a two-phase algorithm. In the first phase, the attacker analyzes the reward function to obtain a set of candidate function estimators. These candidate estimators correspond to all possible local and global maxima of the reward function. We propose novel techniques based on multi-kernel learners to approximate the landscape of local and global maxima and use non-convex constraint solvers to estimate these critical points with bounded error. Note that such analysis of the landscape of critical points can be carried over without having any access to input-output observations from the unknown DNN, and hence we refer to this phase as the offline phase. In the second phase, also called the online phase, the attacker uses the input-output observations to shortlist the number of candidate function estimators. We provide theoretical guarantees on the estimation error between the unknown DNN and the estimated one, for both around the observed input-output data and away from this data. We also show numerical examples to illustrate the efficiency of the proposed algorithm.

\section{Problem Statement}
\subsection{Notations}  
Let $\mathbb{R}$ and $\mathbb{N}$ denote the set of real numbers and natural numbers, respectively. We use $\bigwedge$ and $\bigvee$ to represent the logical AND and logical OR operators, respectively. For a vector $x \in \mathbb{R}^n$, we use $\Vert x\Vert_2$ to denote the Euclidean norm of $x$ and $x_i$ to denote the $i$th element of $x$. Given the two vectors $x \in \mathbb{R}^{n_x}$ and $y \in \mathbb{R}^{n_y}$, we use $(x,y)$ to denote the vector $(x^T, y^T) \in \mathbb{R}^{n_x + n_y}$. Finally, we denote by $|\Gamma|$ the cardinality of a set $\Gamma$ and by $\mathbb{B}_r(c)$ the two-norm ball centered at $c$ with radius $r$, i.e., $\mathbb{B}_r(c) = \{ x \in \mathbb{R}^n | \; \Vert x - c\Vert_2 \le r\}$.

\subsection{System Dynamics}  
We consider non-linear, stochastic, discrete-time dynamic systems of the form:
\begin{align}
\label{eq:dynm}
 \mathbb{P} (s^{(k+1)} \mid s^{(k)},a^{(k)})  = f_{d}(s^{(k)},a^{(k)}), 
 %+ g(s^{(k)},a^{(k)}) 
\end{align}
where $s^{(k)} \in S \subset \mathbb{R}^{q}$ is the state of the system at time $k \in \mathbb{N}$ and $a^{(k)} \in A \subset \mathbb{R}^{m}$ is the action at time $k \in \mathbb{N}$. We assume the dynamical system is controlled by a deep neural network-based policy $\pi_{{\theta}^{*}} {:S \longrightarrow A}$ which is a memory-less, state-feedback, non-linear controller. An $F$-layer NN is specified by composing $F$ layer functions (or just layers). A layer $\omega$ with $\mathfrak{i}_\omega$ inputs and $\mathfrak{o}_\omega$ outputs is specified by a weight matrix $W^{\omega} \in \mathbb{R}^{\mathfrak{o}_\omega \times \mathfrak{i}_\omega}$ and a bias vector $b^{\omega} \in \mathbb{R}^{\mathfrak{o}_\omega}$ as follows:
% The dynamics comprise of $f_{d}$ where $f_{d}$ is the actual model dynamics.
% The state $s^{(k+1)}$ is drawn from the probability distribution defined by $f(s^{(k)},a^{(k)})$. 
% f is kernel of the distribution, once fix input f(s, a) I get a distribution
% s itself is a function, s(k) is the state
\begin{equation}
    \label{eq:layer_fnc}
    L_{\theta^{\omega}}: z \mapsto \phi( W^{\omega} z + b^{\omega}), 
\end{equation}
where $\phi$ is a nonlinear function, $z$ is the input to the NN, and $\theta^{\omega} \triangleq (W^{\omega}, b^{\omega})$ for brevity. Thus, an $F$-layer ReLU NN is specified by $F$ layer functions $\{L_{\theta^{\omega}} : \omega = 1, \dots, F\}$ whose input and output dimensions are composable: that is, they satisfy $\mathfrak{i}_{\omega} = \mathfrak{o}_{\omega-1}$, $\omega = 2, \dots, F$. Specifically:
\begin{equation}
	\pi_\theta(s) = (L_{\theta^{F}} \circ L_{\theta^{F-1}} \circ \dots \circ L_{\theta^{1}})(s),
\end{equation}
where we index a ReLU NN function with a list of parameters $\theta \triangleq (\theta^{1}, \dots, \theta^{F})$. %As a common practice, we allow the output layer $ L_{\theta^{F}}$ to omit the $\max$ function.

%The parameters 

%The RL agent using $\pi_{\boldsymbol{\theta}^{*}}$  takes the current system state, $s$ as an input and produces a control action, $ a $, i.e. $ a^{(k)}=\pi_{\boldsymbol{\theta}^{*}}(s^{(k)})$.

We assume a reinforcement learning algorithm was used to train the policy and obtain the optimal parameters $\theta^*$ that maximizes the discounted, average reward, i.e.:
% \vspace{-1mm}
\begin{align}
  \theta^{*} = \argmax_{\theta \in \Theta} \; J(\theta) = 
  \argmax_{\theta \in \Theta} \; %\mathbb{E}_{s^{(k+1)} \sim f_{d}(s^{(k)}, a^{(k))}}
  \underbrace{\mathbb{E} \sum_{k=0}^{H} \gamma^{k}r\left(s^{(k)},\pi_\theta(s^{(k)}) \right)}_{J(\theta)},
  \label{eq:theta_star}
\end{align}
%
% \r{
% Changes in J due to error in reward:
% \begin{align*}
%   \textbf{Before: }{J(\theta)}=&\mathbb{E} \sum_{k=0}^{H} \gamma^{k}r\left(s^{(k)},\pi_\theta(s^{(k)}) \right)\\
%   \textbf{Now: } J(\theta) =& \mathbb{E} \sum_{k=0}^{H} \gamma^{k}\Big(r\big(s^{(k)},\pi_\theta(s^{(k)}) \big) \r{+ R}\Big) \\
%   =& \mathbb{E} \sum_{k=0}^{H} \gamma^{k}r\big(s^{(k)},\pi_\theta(s^{(k)}) \big)+ \sum_{k=0}^{H} \gamma^{k}\r{R} \\
%     =& \mathbb{E} \sum_{k=0}^{H} \gamma^{k}r\big(s^{(k)},\pi_\theta(s^{(k)}) \big)+ \underbrace{\frac{1-\gamma^{(H+1)}}{1-\gamma}R}_{newerror}
% \end{align*}
% }
%Law of large : numbers avg of samples approaches expected value.
%
where $\Theta \subset \mathbb{R}^{\mathfrak{o}_1 \times (\mathfrak{i}_1 + 1)} \times \ldots \times \mathbb{R}^{\mathfrak{o}_F \times (\mathfrak{i}_F + 1)} \subseteq \mathbb{R}^l$ is a bounded set
%, $\mathfrak{\nu}_{1}, \ldots, \mathfrak{\nu}_{F}$ are the number of parameters inside each layer of the deep neural network policy, 
with
$l = ({\mathfrak{o}_1 \times (\mathfrak{i}_1 + 1)}) + \ldots + ({\mathfrak{o}_F \times (\mathfrak{i}_F + 1)})$ is the total number of parameters in the neural network, the expectation $\mathbb{E}$ is taken over the trajectories of the dynamical system~\eqref{eq:dynm}, $r:S \times A \rightarrow \mathbb{R}$ is the reward function, and $\gamma \in [0, 1)$ is the discounting factor.

\subsection{Threat Model and Problem Definition} 
We assume the role of an attacker who is interested in constructing an estimate $\hat{\theta} \in \Theta$ of the NN policy parameters $\theta^*$ %${\pi}_{\hat{\theta}}$ of the policy $\pi_{{\theta}^{*}}$ 
such that:
% \vspace{-1mm}
\begin{align}
\label{eq:attack_obj}
 \max_{s \in S} \Vert {\pi}_{\hat{\theta}} (s) - \pi_{\theta^{*}}(s) \Vert_2 \le \psi,
\end{align}
for any attacker-selected accuracy $\psi \in \mathbb{R}$ with $\psi > 0$.

We assume the attacker can observe the input-output behavior of the unknown policy $\pi_{{\theta}^{*}}$, i.e.,  $\mathcal{D} = \left\{(s^{(k)}, \pi_{{\theta}^{*}}(s^{(k)})\right\}$ for time step $k = 0, \ldots, N$.
We emphasize that we place no requirements on the quantity and coverage of the observations $\mathcal{D}$. That is important from the attacker's point-of-view since the attacker can not observe the system's behavior in all states. Nevertheless, the attacker's objective in~\eqref{eq:attack_obj} insists on achieving a uniform, attacker-selected error bound across the entire state space. It is direct to show that the requirement in~\eqref{eq:attack_obj} can not be achieved using the observations $\mathcal{D}$, which necessitates the need for additional information. In particular, we assume the attacker has access to a noisy version of the objective function $J$ called $\overline{J}$ where:
  % \vspace{-0.5mm}
\begin{align}
    \max_\theta \Vert \overline{J}(\theta) - J(\theta) \Vert = \max_\theta \Vert m(\theta) \Vert \le \overline{m} ,
    \label{eq:bound_m}
    \vspace{-0.5mm}
\end{align}
where $\overline{m}$ is assumed to be a known upper bound. Such an unknown error function $m(\theta)$ captures the attacker's imperfect information about the system dynamics $f_d$, the reward function $r$, the initial conditions and the approximation of the expectation $\mathbb{E}$ that were used during the training of the policy $\pi_{{\theta}^{*}}$. Hence, whenever the attacker tries to evaluate $J(\theta)$, the attacker will observe an incorrect value of $\overline{J}$ instead, as captured by the unknown error function $m(\theta)$.

\section{Offline Phase: Approximating the Landscape of Critical Points}
To find an estimate ${\pi}_{\hat{\theta}}$ that satisfies~\eqref{eq:attack_obj}, we propose a two-phase algorithm. In the first phase, the attacker uses the imperfect information about $J(\theta)$
%side channel information (e.g., the knowledge of $f_d$, $r$, etc.) 
to compute several candidate estimates, namely ${\pi}_{\hat{\theta}_1}, {\pi}_{\hat{\theta}_2}, \ldots$ without the use of any data collected from the unknown policy. To that end, we first recall that the unknown policy $\pi_{{\theta}^*}$ is the solution to the non-convex optimization problem defined in~\eqref{eq:theta_star}. Gradient-descent-based reinforcement learning algorithms terminate when they reach one global or local maxima. Since the non-convex problem in~\eqref{eq:theta_star} may have several local/global maxima, our objective is to enumerate all these critical points and thus obtain one candidate estimate ${\pi}_{\hat{\theta}}$ for each of the identified local/global maxima. 
% construct one candidate estimates $\widehat{\pi}_{\tilde{\theta}}$ for each of the identified local/global maxima. 
% $\widehat{\pi}_{\tilde{\theta}_1}, \widehat{\pi}_{\tilde{\theta}_2}, \ldots$ 

Formally, we define $g$ as the derivative of the objective function $J$ with respect to the neural network parameters $\theta$, i.e.:
  \vspace{-1mm}
\begin{align}
    g(\theta) = \frac{\text{d}}{\text{d} \theta} J(\theta).
    \label{eq:actual_g}
\end{align}
Indeed, critical points where $g(\theta) = 0$ are the candidate values for the unknown $\theta^*$. Nevertheless, the attacker has only access to an imperfect version of $J$ and hence we define:
\begin{align}
    \overline{g}(\theta) = \frac{\text{d}}{\text{d} \theta} \overline{J}(\theta) = \frac{\text{d}}{\text{d} \theta} (J(\theta) + m(\theta)).
    \label{eq:obs_g}
\end{align}

Our objective is to construct an approximate for $g$ using the information of $\overline{g}$ and isolate candidates for the critical points of $g$. In particular, we will approximate $g$ using multi-kernel regression algorithms. Multi-kernel-based estimators provide closed-form expressions that can be exploited for our analysis. To that end, we proceed as below.

\subsection{Step 1: Use multi-kernel regression to estimate $g(\theta)$:} 

Using a grid of size $\eta$, we uniformly sample from the space of parameters $\Theta$ and we denote by $\tilde{\theta}^{(1)}, \ldots, \tilde{\theta}^{(M)}$, the sampled neural network parameters. For each sample, the attacker aims to evaluate $J(\tilde{\theta}^{(i)})$ using the imperfect knowledge of the dynamics and the reward function and obtains $\overline{J}(\tilde{\theta}^{(i)})$ which can be used to approximate the derivative $g$ by using the information of $\overline{g}$. 
% \vspace{3mm}
Specifically $\overline{g} = (\overline{g}_1, \ldots, \overline{g}_l)$ will be approximated by $\widetilde{g}_j$ as:
% \vspace{3mm}
\begin{align}
\widetilde{g}_j(\tilde{\theta}) &= \frac{\overline{J}(\tilde{\theta} + c_j) - \overline{J}(\tilde{\theta})}{c}
\nonumber \\
& =\frac{J(\tilde{\theta}+c_j) + m(\tilde{\theta}+ c_j) - J(\tilde{\theta}) - m(\tilde{\theta})}{c},
\label{eq:gtilda}
\end{align}
where $j \in \{1,...,l\}$ and $c_j \in \mathbb{R}^l$ is defined as a vector of all zeros except for the $j$-th element, which is equal to $c$. %Thanks to the side-information knowledge about $f_d$, $r$, and $\gamma$, one can evaluate $J$ at any given value of $\theta$.
The collected samples $\{(\tilde{\theta}^{(i)},{\widetilde{g}_{j}}(\tilde{\theta}^{(i)}))\}_{i = 1}^M$ are used to learn a set of Multi-Kernel Learner (MKL)-based function estimators $\widehat{g}_j(\theta)$ of the form:
\vspace{-2mm}
\begin{align}
\widehat{g}_j(\theta) = \alpha_{j}^T k(\theta) && j \in \{1,...,l\},
\end{align}
where $k(\theta) = \left(\kappa(\theta, \tilde{\theta}^{(1)}), \ldots, \kappa(\theta, \tilde{\theta}^{(M)}) \right)$ is the kernel vector\footnote{In practice, we use feature-based MKL which ensures that the dimension of the kernel vector $k$ does not increase with the number of samples, see~\cite{shen2019random}.} with $\kappa(\theta,\tilde{\theta}):\mathbb{R}^l \times \mathbb{R}^l \rightarrow \mathbb{R}$ is a symmetric positive semidefinite basis (so-termed kernel) function, which measures the similarity between $\theta$ and $\tilde{\theta}$. With some abuse of notation, the vector $\alpha_j \in \mathbb{R}^M$ is the solution to the optimization problem:
\vspace{-2mm}
\begin{align}
    \label{eq:mkl_optimization}
  \alpha_j =\argmin_{\alpha_j \in \mathbb{R}^M} \sum_{i = 1}^M \|\widehat{g}_j(\theta) - \widetilde{g}_j(\tilde{\theta}^{(i)})\|_2 && j \in \{1,...,l\}.
  \vspace{-1mm}
\end{align}
Solving~\eqref{eq:mkl_optimization} has been widely studied in the literature; see for example~\cite {shen2019random}. We can bound the error between $g$ and its estimate $\widehat{g}$ as follows.

\vspace{-0.7mm}
\begin{proposition} \label{prop:1}
Under the following assumptions:
\begin{enumerate}
    \item $\exists R \in \mathbb{R} \; \text{ such that } \; \Vert r(s,a)\Vert_2 \le R \quad \forall (s,a)\in S \times A,$
    \item $\exists G \in \mathbb{R} \; \text{ such that } \; \Vert \frac{\partial}{\partial \theta} \pi_{\theta}(s) \Vert_2 \le G \quad \forall s\in S,$
    \item $\exists L \in \mathbb{R} \; \text{ such that } \; \Vert \frac{\partial^2}{\partial \theta^2} \pi_{\theta}(s) \Vert_2 \leq L \quad \forall s\in S,$
\end{enumerate}

then the component-wise maximum estimation error is bounded as:
\begin{align}
\max_{\theta \in \Theta} \|g_{j}(\theta) - \widehat{g}_{j}(\theta)\|_2 \le \underbrace{c {L_{g_{j}}} + \frac{2\overline{m}}{c} + \zeta_{j} + \eta ({L_{\widehat{g}_{j}}} + {L_{\tilde{g}_{j}}})}_{e_{j}},
\label{eq:component_e1}
\end{align}

 where $j \in \{1, \ldots, l\}$, and $L_{{g}_{j}}$, $L_{\widehat{g}_{j}}$ and $L_{\tilde{g}_{j}}$ are the Lipschitz constants of $g_{j}$, $\widehat{g}_{j}$ and $\tilde{g}_{j}$ respectively, while $\zeta_j$ is the maximum sample error at the $j$th component, defined as $\zeta_j = \max_{i \in \{1, \ldots, M\}} \Vert \widehat{g}_j(\tilde{\theta}^{(i)}) - \tilde{g}_j(\tilde{\theta}^{(i)}) \Vert_2$. 
\end{proposition}
% while $\zeta$ is the maximum sample error defined as $\zeta = \Vert (\zeta_1, \ldots, \zeta_l) \Vert$ and $\zeta_j = \max_{i \in \{1, \ldots, M\}} \Vert \widehat{g}_j(\tilde{\theta}^{(i)}) - \tilde{g}_j(\tilde{\theta}^{(i)}) \Vert_2$. 
%  where $L_{{g}}$, $L_{\widehat{g}}$ and $L_{\tilde{g}}$ are the Lipschitz constants of $g$, $\widehat{g}$ and $\tilde{g}$, respectively

%

Under the assumptions in Proposition~\ref{prop:1}, the Lipschitz constant $L_g$ (and hence the Lipschitz constant  $L_{g_j}$ of the component-wise function $g_j$) can be bounded as $L_g \le \frac{H^2 G^2 R^2 + L^2 R^2}{1-\gamma^4}$. The Lipschitz constant $L_{\widehat{g}}$ can be directly computed from the knowledge of the Lipschitz constants of the individual kernel function $\kappa$ and the vectors $\alpha_1, \ldots, \alpha_l$. Finally, the Lipschitz constant $L_{\tilde{g}_j}$ can be computed from the samples $\{(\tilde{\theta}^{(i)},\widetilde{g}_j(\tilde{\theta}^{(i)})\}_{i = 1}^M$ that were used to define the function $\widetilde{g}_j$ using methods from optimization theory~\cite{wood1996estimation,fainekos2009robustness} since we have samples of the function, not the analytical form of the function.
%  To see the relation between the Lipschitz constant of any function $g$ and its individual components $g_j$, please read the Appendix.

%
\begin{remark} \label{remark:1}
The assumptions in Proposition~\ref{prop:1} are standard assumptions in the reinforcement learning literature, see for example~\cite{shen2019hessian}, \cite{papini2018stochastic}, and the references within.
\end{remark}
%

% \vspace{+5mm}
\begin{algorithm}[!b]
\caption{Offline Phase}\label{alg:offline}
   \textbf{Input:} {Objective function $J$, Dynamics $f_d$, Kernel vector $k(\theta)$, Neural network architecture/parameters bound $\Theta$, and attacker-threshold $\psi$.
   }
  
   \textbf{Output:} {Candidate policy esti$\{{\pi}_{\hat{\theta}^{(1)}}, \ldots, {\pi}_{\hat{\theta}^{(r)}}\}$
   }
  \begin{algorithmic}[1]
  \STATE Set $b \le \frac{\psi}{G}$
  \STATE Sample uniformly from the policy parameter space $\Theta$ along a grid of size $\eta$ to obtain $\tilde{\theta}^{(1)}, \ldots, \tilde{\theta}^{(M)}$\;
  
  \FOR{$j \in \{1,...,l\}$}
  
    \FOR{$\tilde{\theta} \in \{\tilde{\theta}^{(1)}, \ldots, \tilde{\theta}^{(M)}\}$}
    
        \STATE Evaluate $\overline{J}$ to obtain:
          \begin{align*}
          \widetilde{g}_j(\tilde{\theta}) & =\frac{\overline{J}(\tilde{\theta}+c_j) - \overline{J}(\tilde{\theta})}{c} %\\
        %J(\theta) &= \mathbb{E} \sum_{k=0}^{H} \gamma^{k}\Big(r\big(s^{(k)},\pi_\theta(s^{(k)}) \big)\Big)
        \end{align*}
    \vspace{-0.5mm}
    \ENDFOR

  \STATE 

  Use the samples $\{(\tilde{\theta}^{(i)},\widetilde{g}_j({\tilde{\theta}}^{(i)})\}_{i = 1}^M$ to train a multi-kernel learner and construct the function estimator $\widehat{g}_j$ as:
    \begin{align*}
        \widehat{g}_j(\theta) = \alpha_{j}^T k(\theta)
    \end{align*}
    \vspace{-1mm}
  \ENDFOR

    \STATE Set \texttt{Finished}:= FALSE and $i := 0$
    \WHILE{\texttt{Finished} = FALSE}
        \STATE $i := i + 1$
        \STATE Use SMT solver to find a feasible solution of:
        \begin{align*} 
            \exists &\hat{\theta}^{(i)} \in \Theta . 
            \\ & \left[
            - \overline{e} \le \widehat{g}(\hat{\theta}^{(i)}) \leq  \overline{e} \bigwedge \left( \bigvee_{j = 2}^{i-1} \Vert \hat{\theta}^{(i)} - \hat{\theta}^{(j)}\Vert \ge b \right)
        \right]
        \end{align*}
    \ENDWHILE
    \STATE Set $r :=$ number of solutions found by the SMT solver
  \RETURN Candidate policy estimates 
  $\{\pi_{\hat{\theta}^{(1)}}, \ldots, \pi_{\hat{\theta}^{(r)}}\}$
%   $\{\widehat{\pi}_{\hat{\theta}^{(1)}}, \ldots, \widehat{\pi}_{\hat{\theta}^{(r)}}\}$
    \end{algorithmic}
\end{algorithm}

\subsection{Step 2: Solve a sequence of feasibility problems to identify critical points:} 
The next step is to use the estimated function $\widehat{g}$ (and the error margin $e_j$ in Proposition~\ref{prop:1}) to identify candidate neural network parameters $\hat{\theta}^{(1)}, \hat{\theta}^{(2)}, \ldots$ that are guaranteed to be close enough to the critical points of $J$. To that end, we solve the following feasibility problem:
\begin{align} 
\exists \hat{\theta}^{(i)} & \in \Theta . \notag 
\\ & 
\left[
- \overline{e} \le \widehat{g}(\hat{\theta}^{(i)}) \leq  \overline{e} \bigwedge \left( \bigvee_{j = 2}^{i-1} \Vert \hat{\theta}^{(i)} - \hat{\theta}^{(j)}\Vert \ge b \right)\right], \notag
\\
\label{eq:smt1}
\end{align}

for $i = 1, 2, \ldots$ until no more solutions can be found. The $-\overline{e} \le \widehat{g}(\hat{\theta}^{(i)}) \leq  \overline{e}$ constraint in~\eqref{eq:smt1} ensures that the identified $\hat{\theta}^{(i)}$ is close to the local/global maxima of $J$. In this constraint, the vector $\overline{e} \in \mathbb{R}^l$ is defined as $\overline{e} = ({e_1},\ldots, {e_j},\ldots,{e_l})$ where $e_j$ is the right-hand side of \eqref{eq:component_e1}.
The second set of constraints $\bigvee_{j = 2}^{i-1} \Vert \hat{\theta}^{(i)} - \hat{\theta}^{(j)}\Vert \ge b$ ensures that the distance between $\hat{\theta}^{(i)}$, at iteration $i$, and those found in the previous iterations $\hat{\theta}^{(1)}, \hat{\theta}^{(2)}, \ldots, \hat{\theta}^{(i-1)}$ is larger than a user-defined parameter $b \in \mathbb{R}$ with $b > 0$.

For each iteration $i = 1, 2, \ldots$, the feasibility problem in~\eqref{eq:smt1} can be efficiently solved using Satisfiability Modulo Theory (SMT) solvers. SMT solvers are capable of solving feasibility problems that can be encoded as first-order logic formulas over the real numbers and has been widely used to solve numerous problems in computer science and control theory literature~\cite{shoukry2017secure,huang2015controller,bahavarnia2020controller,barrett2018satisfiability}. Examples are dReal which can handle analytic functions~\cite{gao2013dreal}, PolyAR and PolyARBerNN which are optimized to handle polynomial functions~\cite{fatnassi2021polyar}\cite{fatnassi2022polyarbernn}, and Z3 which is a generic SMT solver~\cite{moura2008z3}. The next result captures the correctness of this iterative procedure.

\begin{proposition}
\label{prop:2}
Consider the set of neural network parameters $\{\hat{\theta}^{(1)}, \ldots, \hat{\theta}^{(r)}\}$ computed by iteratively solving the feasibility problem~\eqref{eq:smt1} for $h = 1, \ldots, r$ until no more solutions can be found, where $b >0$ is any user-defined threshold. Under the assumptions in Proposition~\ref{prop:1}, the following holds:
% \vspace{-1mm}
\begin{align}
    \exists \hat{\theta}^{(i^*)} \in \{\hat{\theta}^{(1)}, \ldots, \hat{\theta}^{(r)}\} \text{ such that } \Vert \theta^* - \hat{\theta}^{(i^*)} \Vert_2 \le b.
    \label{eq:prop2}
\end{align}
\end{proposition}

Algorithm~\ref{alg:offline} summarizes the steps in the offline phase, and the following Theorem represents the correctness guarantees of the candidate policies obtained from the offline phase.

\begin{theorem}
\label{th:offline}
Consider an unknown neural network controller $\pi_{{\theta}^*}$ with $\theta^*$ defined as in~\eqref{eq:theta_star}. For any attacker-specified threshold $\psi > 0$, and parameter  $b \le \frac{\psi}{G}$:
\begin{align}
    \exists {\pi}_{\hat{\theta}^{i^*}} \in \{{\pi}_{\hat{\theta}^{(1)}}, & \ldots, {\pi}_{\hat{\theta}^{(r)}}\} \text{ s.t. } \nonumber
    \\
    &
    \max_{s \in S} \Vert \pi_{\theta^{*}}(s) - {\pi}_{\hat{\theta}^{(i^*)}} (s) \Vert_2 \le \psi,
\end{align}
where $\{{\pi}_{\hat{\theta}^{(1)}}, \ldots, {\pi}_{\hat{\theta}^{(r)}}\}$ is the set of polices computed by Algorithm~\ref{alg:offline}.
\end{theorem}
% \vspace{-1mm}
In other words, Theorem~\ref{th:offline} ensures that one of the polices computed by Algorithm~\ref{alg:offline} is guaranteed to satisfy the attacker-provided threshold. Nevertheless, Algorithm~\ref{alg:offline} is not guaranteed to identify which of the computed polices is the closest one to the unknown policy. This motivates the online phase of our algorithm discussed in the next section.

\begin{algorithm} [!h]
\caption{Online Phase}\label{alg:online}
\textbf{Input:} {Set of the shortlisted candidate policy estimates $\pi_\text{Str}$, set of the discarded policy estimates $\pi_\text{Dis}$, and attacker-threshold $\psi$. \\
Before the first run of the algorithm, these two sets are initialized to $\pi_\text{Str} = \{\pi_{\hat{\theta}^{(1)}}, \ldots, \pi_{\hat{\theta}^{(r)}} \}$ and $\pi_\text{Dis} = \phi$.}

\textbf{Output:} {An updated set of shortlisted policy estimate(s) $\pi_\text{Str}$ and an updated set of discarded policy estimates $\pi_\text{Dis}$}
    % \vspace{+4mm}
    \begin{algorithmic}[1]
    
    \STATE Obtain the samples $({s}^{(k)},\pi_{\theta^{*}}({s}^{(k)}))$ as they become available.
    \FOR{$\pi_{\hat{\theta}} \in \pi_\text{Str}$}
    {
    \STATE Evaluate the neural network $\pi_{\hat{\theta}}$ to get the estimated control action $\pi_{\hat{\theta}}({s}^{(k)})$
    % \STATE Compare the difference between the optimal and estimated policy outputs as below: 
    \IF{$ \| {\pi_{\hat{\theta}}}({s}^{(k)}) - \pi_{{{\theta}}^{*}}({s}^{(k)}) \|_2 > \psi $:} 
    {\item $\pi_\text{Dis} = \pi_\text{Dis} \bigcup \{\pi_{\hat{\theta}}({s}^{(k)})\}$
    \item $\pi_\text{Str} = \pi_\text{Str} / \{\pi_{\hat{\theta}}({s}^{(k)})\} $
    }
    \ENDIF
     }
     \ENDFOR
     
    \STATE $q = |\pi_\text{Str}| $ is number of policies satisfying the threshold

    \RETURN $\pi_\text{Str}$ and $\pi_\text{Dis}$
    \end{algorithmic}
%   \vspace{+4mm}
\end{algorithm}

\section{Online Phase: Policy Estimate Selection}
In this phase, we shortlist the set of final policy estimates $\pi_\text{Str}$ from all the candidate policies of the offline phase $\{\pi_{\hat{\theta}^{(1)}}, \ldots, \pi_{\hat{\theta}^{(r)}}\}$. This is done by comparing the outputs of the policies with those observed from the unknown policy. The selection is done based on the user-specified threshold $\psi$ in Theorem \ref{th:offline}. That is, once a new pair $({s}^{(k)},\pi_{\theta^{*}}({s}^{(k)}))$ is observed at time $k$, we evaluate all the candidate policies to obtain $\pi_{\hat{\theta}^{(i)}}({s}^{(k)})$, for $i \in \{1, \ldots, r\}$. Next, we compare these outputs to the observed $\pi_{\theta^{*}}({s}^{(k)})$ and only leave those policies that produce actions within the $\psi$ threshold. Algorithm~\ref{alg:online} summarizes the steps in the online phase. It is direct to show that Theorem 3 still holds for the shortlisted policies $\pi_\text{Str}$ as captured by the Corollary~\ref{th:online}.
% \vspace{-10mm}
% \vspace{-10mm}
\begin{corollary}
\label{th:online}
Consider an unknown neural network controller $\pi_{{\theta}^*}$ with $\theta^*$ defined as in~\eqref{eq:theta_star}. For any attacker-specified threshold $\psi > 0$:
\vspace{-1mm}
\begin{align}
\vspace{-1mm}
    \exists {\pi}_{\hat{\theta}^{i^*}} \in \pi_{\text{Str}} \text{ s.t. } 
    \max_{s \in S} \Vert \pi_{\theta^{*}}(s) - {\pi}_{\hat{\theta}^{(i^*)}} (s) \Vert_2 \le \psi,
    \vspace{-3mm}
\end{align}
\vspace{-1.6mm}
where $\pi_{\text{Str}}$ is the set of policies computed by Algorithm~\ref{alg:online}.
\end{corollary}

We note that the reduction in the number of shortlisted policies (compared to those produced in the offline phase) depends on the quality and quantity of the collected data. Nevertheless, as we discuss in the next section, even very few samples can be enough to reduce the number of shortlisted policies to a small number.

\section{Numerical Experimentation}
\subsection{Environment Setup 1} 
We use the widely used platform, OpenAI Gym utilizing Box2D Physics engine. It is a toolkit for reinforcement learning research \cite{brockman2016openai}, which provides various dynamical systems and control tasks, amongst others, for testing. The exact dynamics are unknown to the attacker. %, however it does use data collected in the offline phase to learn the landscape of critical points. 

\begin{figure}[h!]
    \centering
\vspace{-5mm}
    \includegraphics[scale=0.183]{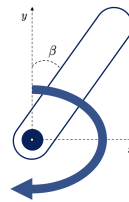}
    \caption{Inverted Pendulum}
    \label{fig:Inv_Pend}
\end{figure}

We use the classic inverted pendulum task as an example, shown in Figure~\ref{fig:Inv_Pend}, with the default parameters of the physical environment in the OpenAI Gym, which results in the following state space dynamics:
\begin{equation}
    \frac{\text{d}}{\text{d} t} \begin{bmatrix}
        x\\
        y\\
        \dot{\beta}
    \end{bmatrix}
    = \begin{bmatrix}
        0 & -1 & 0\\
        1 & 0 & 0\\
        0 & \frac{3gr}{2l} & 0
    \end{bmatrix}
    \begin{bmatrix}
        x\\
        y\\
        \dot{\beta}
    \end{bmatrix}
    +
    \begin{bmatrix}
    0\\
    0\\
    \frac{3}{ml^2}
    \end{bmatrix}
    u, 
    \vspace{-0.9mm}
\end{equation}
where the state vector $(x,y,\dot{\beta})$ represents the position and the angular velocity of the pendulum, $m$ is the mass and $l$ is the length of the pendulum, $gr$ is the acceleration of gravity, and $t$ is the time step. We used the following reward function:
 \vspace{-1.3mm}
\begin{equation}
    r = -(\beta^2 + 0.1\dot{\beta}^2 + 0.001u^2), \vspace{-1mm}
\end{equation}
to train a neural network controller using actor-critic PPO with $F = 3$ and $l = 11$. Indeed the learned $\theta^*$ is unknown to the attacker.

\subsection{Numerical Results 1} 

\begin{figure}[!h]
 \centering  \includegraphics[width=0.48\textwidth]{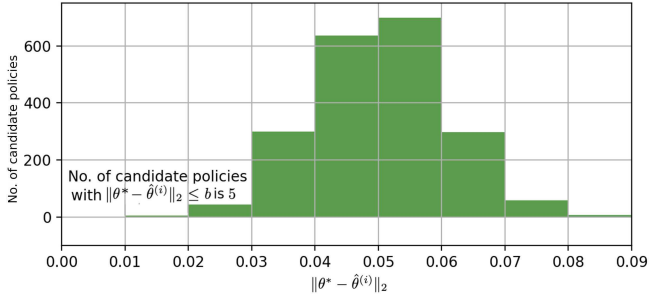}
    \caption{Histogram of the error  $\Vert \theta^* - \hat{\theta}^{(i)} \Vert_2$ across all identified candidate policies by Algorithm~\ref{alg:offline}.}
    \label{fig:pend1}
\end{figure}

\begin{figure} [h!]
    % \centering
    \includegraphics[scale=0.535]{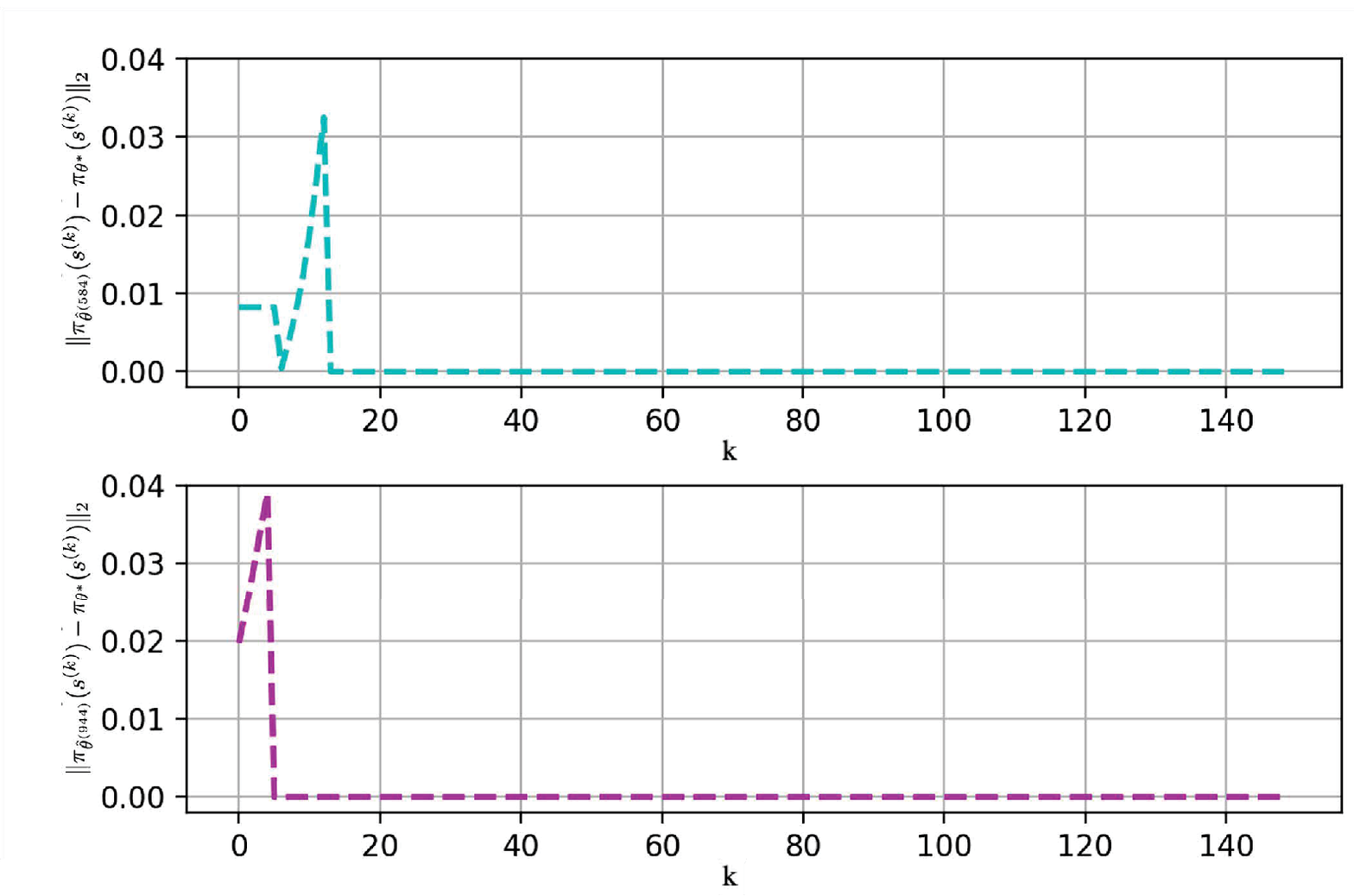}
    \caption{Error $\| {\pi_{\hat{\theta}}}({s}^{(k)}) - \pi_{{{\theta}}^{*}}({s}^{(k)}) \|_2$ of all candidate policies ${\pi_{\hat{\theta}}} \in  \pi_\text{Str}$, where $\psi = 0.04, r=2047, q=2,\text{ and } \pi_{\text{Str}} = \{\pi_{\hat{\theta}^{584}}, \pi_{\hat{\theta}^{944}}\}$, generated by Algorithm $\ref{alg:online}$.}
     \label{fig:traj_err}
\end{figure}

\begin{figure}[h!]
    \centering
    \includegraphics[scale=0.29]{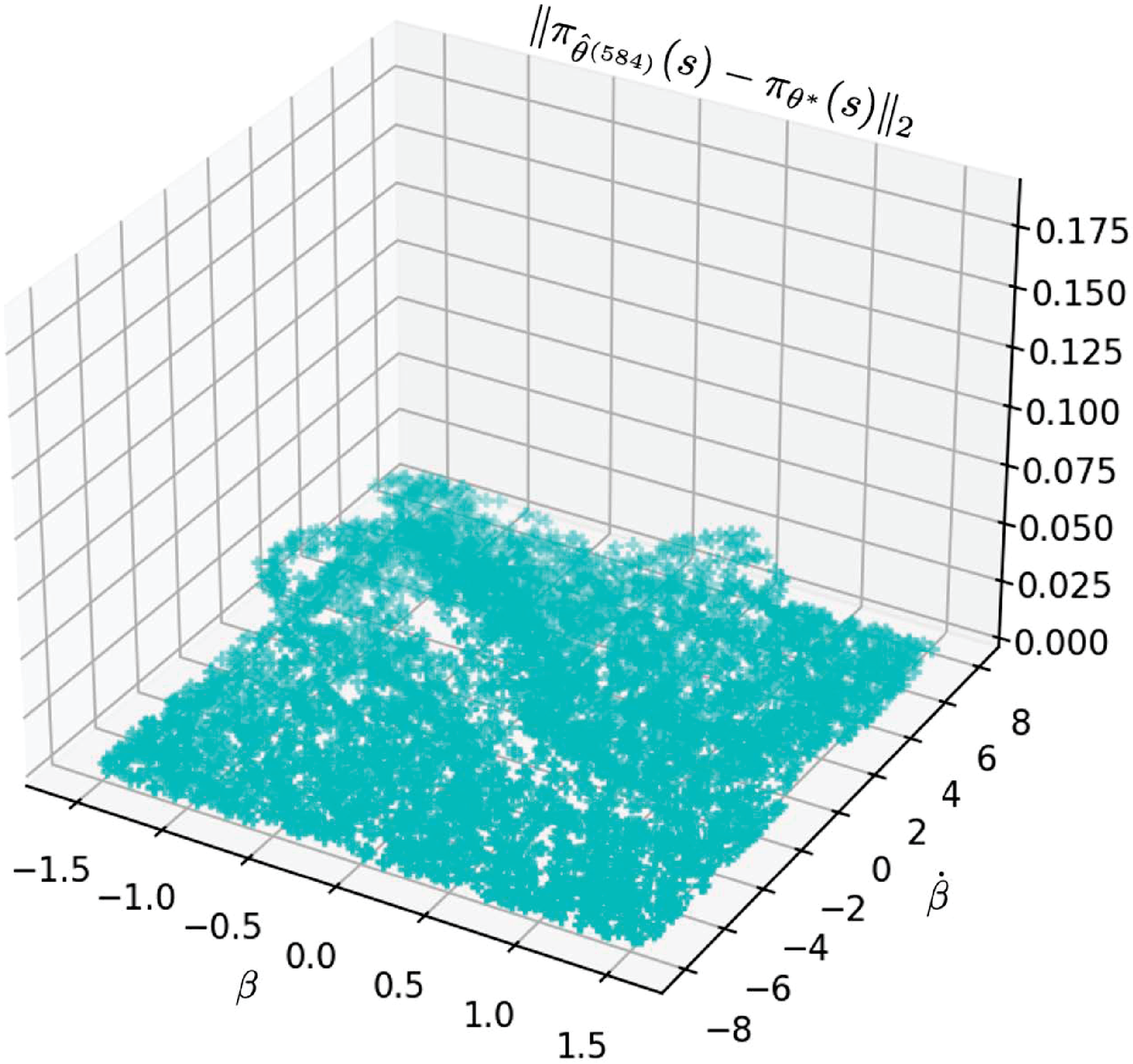}
    \includegraphics[scale=0.29]{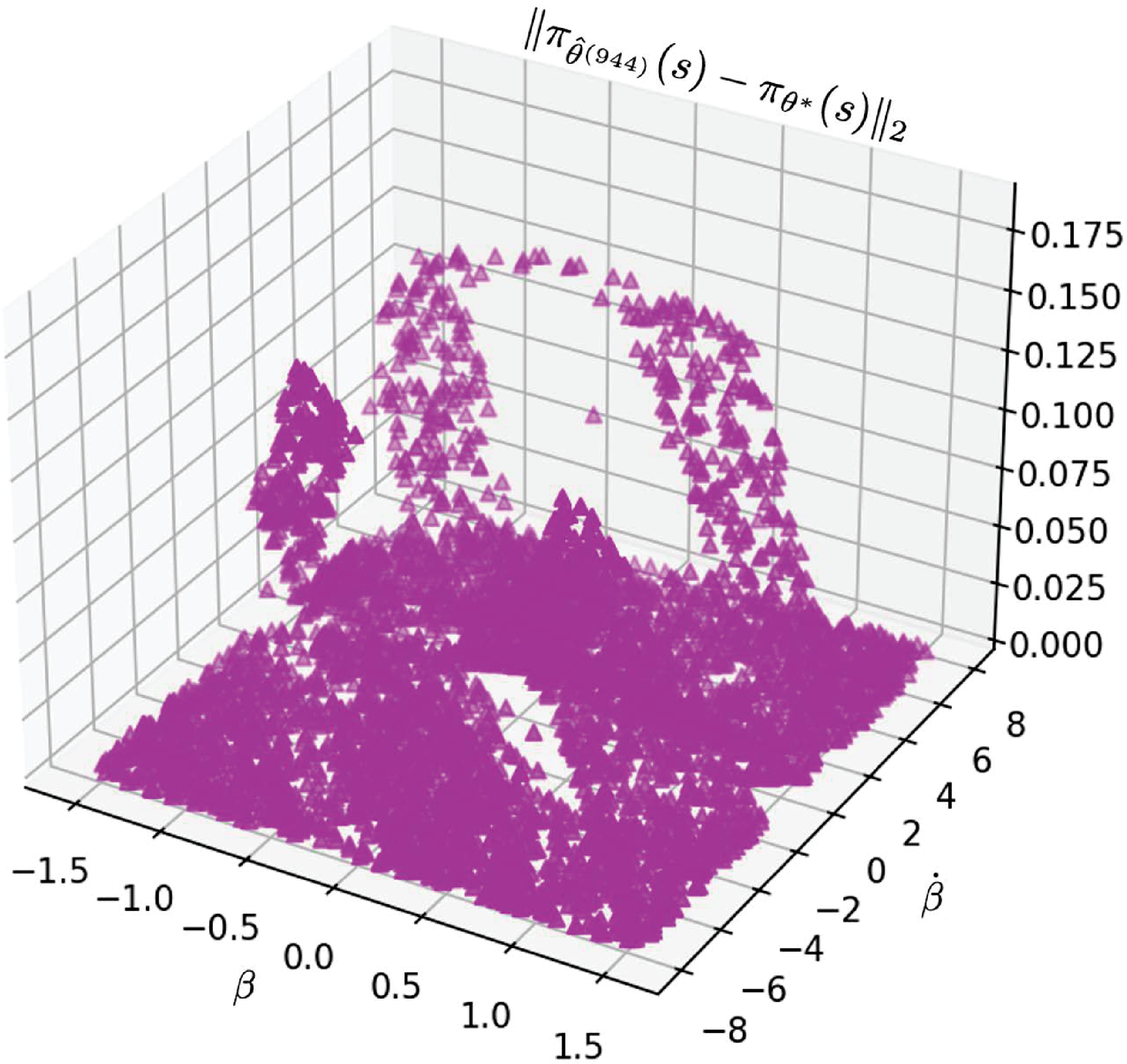}
    \caption{Error $\| {\pi_{\hat{\theta}}}({s}) - \pi_{{{\theta}}^{*}}({s}) \|_2$ , over uniform samples across the state space of ${\pi_{\hat{\theta}}} \in  \pi_\text{Str}$ generated by Algorithm \ref{alg:online}.}
    \label{fig:err}
\end{figure}

We first run our offline phase and collect all the candidate policy estimates $\{\pi_{\hat{\theta}^{(1)}}, \ldots, \pi_{\hat{\theta}^{(r)}}\}$. %As mentioned, we have a 3-layer NN with a total of $l=11$ parameters so $\theta \in \mathbb{R}^{11}$. 
We use $c = 0.05$ and we set $\psi = 0.04$. We estimated the parameters $G$, $L_g$, $L_{\widehat{g}}$ and $L_{\tilde{g}}$ from data. Using these parameters, we set $b = 0.02$ and by running Algorithm 1, we obtained $r=2047$ candidate policies. Figure~\ref{fig:pend1} shows the histogram of the error $\Vert \theta^* - \hat{\theta}^{(i)} \Vert_2$ across all identified candidate policies. As can be observed, there exist $5$ candidate NN parameters that are within the bound $b$ promised by Proposition \ref{prop:2}.
\\
Next, we ran Algorithm~\ref{alg:online} to get the final set of candidate policy estimates $\pi_\text{Str}$.
To illustrate the power of Algorithm~\ref{alg:online} in terms of its ability to shortlist the candidate policies, we ran Algorithm~\ref{alg:online} for $25$ times from different initial conditions of the system. In all runs, we observed $N = 150$ samples from the unknown RL agent. In all runs of Algorithm~\ref{alg:online}, it was capable of eliminating $99\%$ of candidate policies. More specifically, the number of shortlisted policies $q = |\pi_\text{Str}|$ varied between $1$ and $13$ with an average of $4.24$ across the 25 runs. Figure \ref{fig:traj_err} shows the results of one of these runs which resulted in $\pi_\text{Str} = \{\pi_{\hat{\theta}^{(584)}}, \pi_{\hat{\theta}^{(944)}}\}$. In particular, Figure \ref{fig:traj_err} shows the error $\| {\pi_{\hat{\theta}^{(i)}}}({s}^{(k)}) - \pi_{{{\theta}}^{*}}({s}^{(k)}) \|_2$ across the trajectory of collected observations for both $\pi_{\hat{\theta}^{(584)}}$ and $\pi_{\hat{\theta}^{(944)}}$. As can be observed from the figure, both candidates maintain an error that is below the user-provided threshold of $\psi=0.04$.
\\ 
In Figure \ref{fig:err}, we randomly sample across the entire state space and plot the error for the two shortlisted candidate policies $\pi_\text{Str}$. As promised by Corollary~\ref{th:online}, at least one of the identified policies has a maximum error that is less than $\psi=0.04$. More specifically the maximum error for ${\pi_{\hat{\theta}^{(584)}}} = 0.037$.

\subsection{Environment Setup 2} 
\begin{figure}[h!]
    \centering   
    \vspace{-2.5mm}
    \includegraphics[scale=0.23]{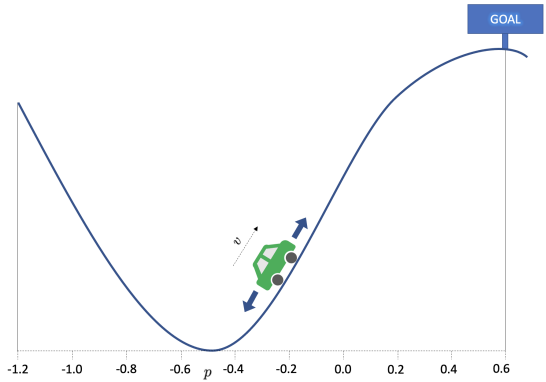}
    \caption{Mountain Car}
    \label{fig:MC}
    \vspace{-1.5mm}
\end{figure}

Next, we used the OpenAI Gym mountain car example where the car is placed stochastically at the bottom of a sinusoidal valley, with the possible actions being acceleration in either direction. The goal is to reach the goal state on top of the right hill, shown in Figure~\ref{fig:MC}. The walls at both the ends are inelastic. The system dynamics follow the following nonlinear dynamics:
\begin{equation}
    \frac{\text{d}}{\text{d} t} \begin{bmatrix}
        p\\
        v\\
    \end{bmatrix}
    = \begin{bmatrix}
        v\\
        0.001u - 0.0025\cos(3p) - 0.0025v^{2}\sin(3p)
    \end{bmatrix}.
\end{equation}

The reward function used for training is:
\begin{align}
    & r = - 0.1u^2  \text{\hspace{0.5mm}for each t not reached goal} \notag
    \\ & r = + 100  \text{\hspace{4mm}reached the goal}
\end{align}
We used actor-critic PPO to train a neural network controller with 3 layers and 13 neurons/layer.
%This was used to train a neural network controller using actor-critic PPO with $F = 3$ and $l = 13$. 
Indeed the learned $\theta^*$ is unknown to the attacker. 

This setup involves non-linear dynamics with multiple local optima, making it difficult to find a globally optimal policy to solve the problem. This setup is known to suffer from several local minima, and the reward function is sparse and delayed, providing limited feedback to the agent during the learning process.

\subsection{Numerical Results 2} 
\begin{figure}[!t]
 \centering
 \vspace{1.8mm}
    \includegraphics[width=0.48\textwidth]{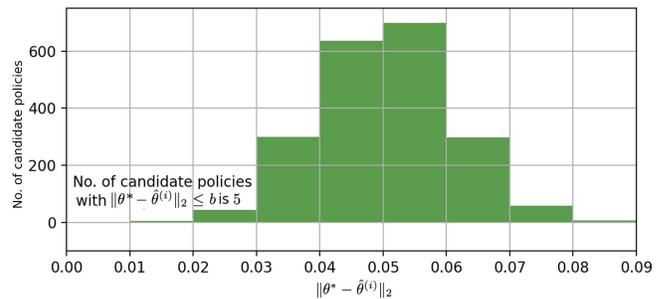}
     \caption{Histogram of the error  $\Vert \theta^* - \hat{\theta}^{(i)} \Vert_2$ across all identified candidate policies by Algorithm~\ref{alg:offline}.}
    \label{fig:car11}
\end{figure}

We first run our offline phase and collect all the candidate policy estimates $\{\pi_{\hat{\theta}^{(1)}}, \ldots, \pi_{\hat{\theta}^{(r)}}\}$. %As mentioned, we have a 3-layer NN with a total of $l=11$ parameters so $\theta \in \mathbb{R}^{11}$. 
We use $c = 0.05$ and we set $\psi = 0.55$. We estimated the parameters $G$, $L_g$, $L_{\widehat{g}}$ and $L_{\tilde{g}}$ from data. Using these parameters, we set $b = 0.37$ and by running Algorithm 1, we obtained $r=9136$ candidate policies. Figure~\ref{fig:car11} shows the histogram of the error $\Vert \theta^* - \hat{\theta}^{(i)} \Vert_2$ across all identified candidate policies. As can be observed, there exist $7$ candidate NN parameters that are within the bound $b$ promised by Proposition \ref{prop:2}.

Next, we ran Algorithm~\ref{alg:online} to get the final set of candidate policy estimates $\pi_\text{Str}$.
%To illustrate the power of Algorithm~\ref{alg:online} in terms of its ability to shortlist the candidate policies, 
We ran Algorithm~\ref{alg:online} for $25$ times from different initial conditions of the system. In all runs, we observed $N = 150$ samples from the unknown RL agent. In all runs of Algorithm~\ref{alg:online}, it could eliminate most candidate policies. More specifically, the number of shortlisted policies $q = |\pi_\text{Str}|$ varied between $2$ and $23$ with an average of $6.13$ across the 25 runs. 

\begin{figure} [h!]
    \centering
    \includegraphics[scale=0.121]{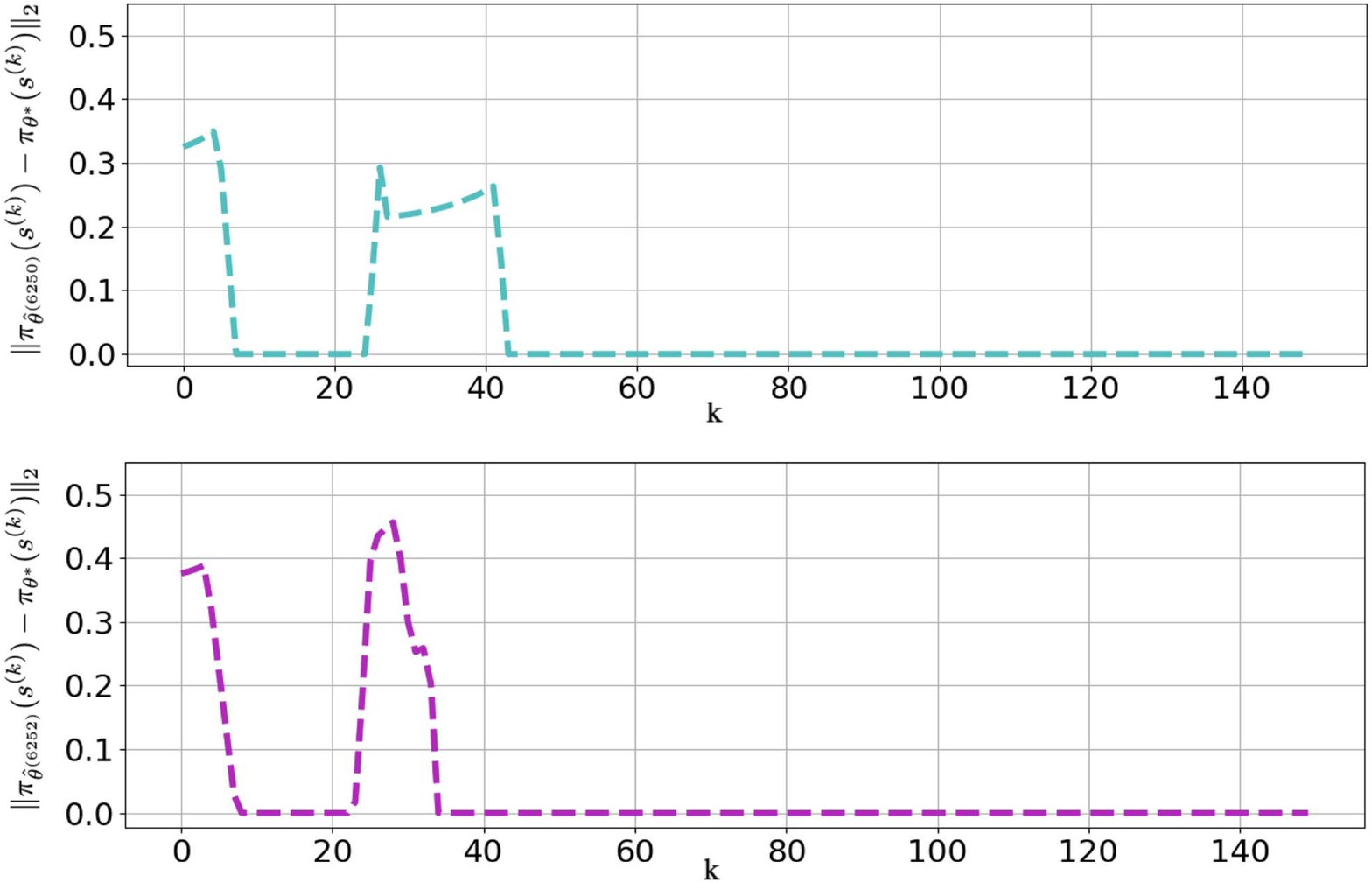}
    \includegraphics[scale=0.1226]{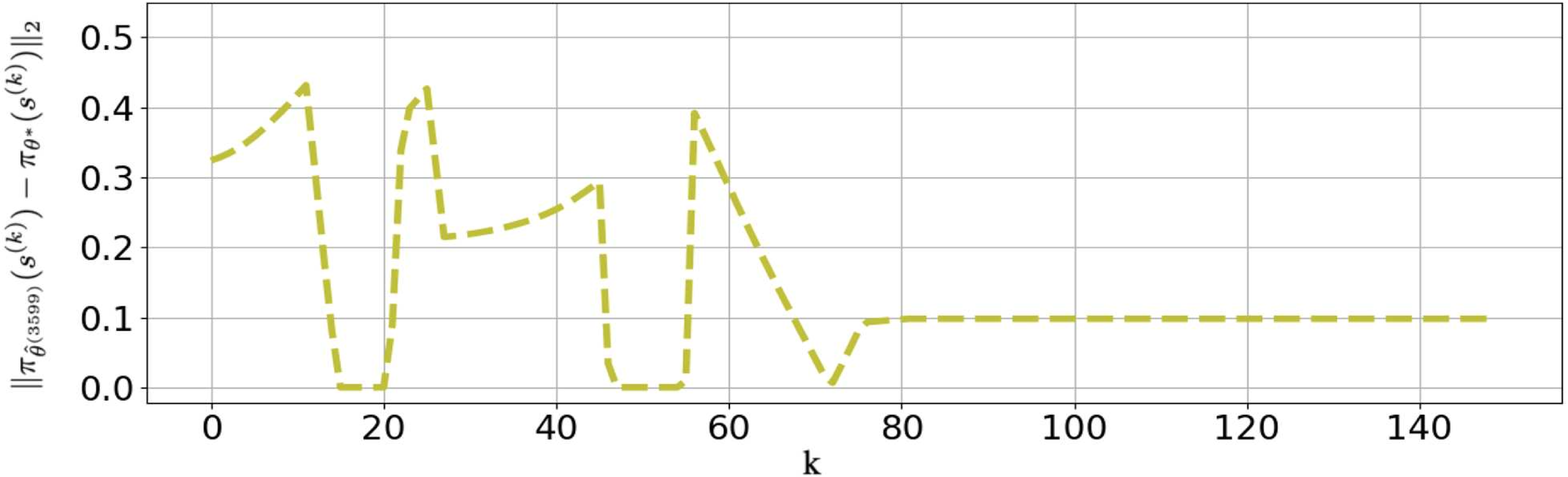}
    \caption{Error $\| {\pi_{\hat{\theta}}}({s}^{(k)}) - \pi_{{{\theta}}^{*}}({s}^{(k)}) \|_2$ of all candidate policies ${\pi_{\hat{\theta}}} \in  \pi_\text{Str}$, where $\psi = 0.55, r = 9136, q=3,\text{ and } \pi_{\text{Str}} = \{\pi_{\hat{\theta}^{6250}}, \pi_{\hat{\theta}^{6252}}, \pi_{\hat{\theta}^{3599}}\}$, generated by Algorithm \ref{alg:online}. 
    }
     \label{fig:traj_err_exp2}
\end{figure}

\begin{figure*}[t!]
\vspace{2.5mm}
    \centering
    \includegraphics[width=0.326\textwidth]{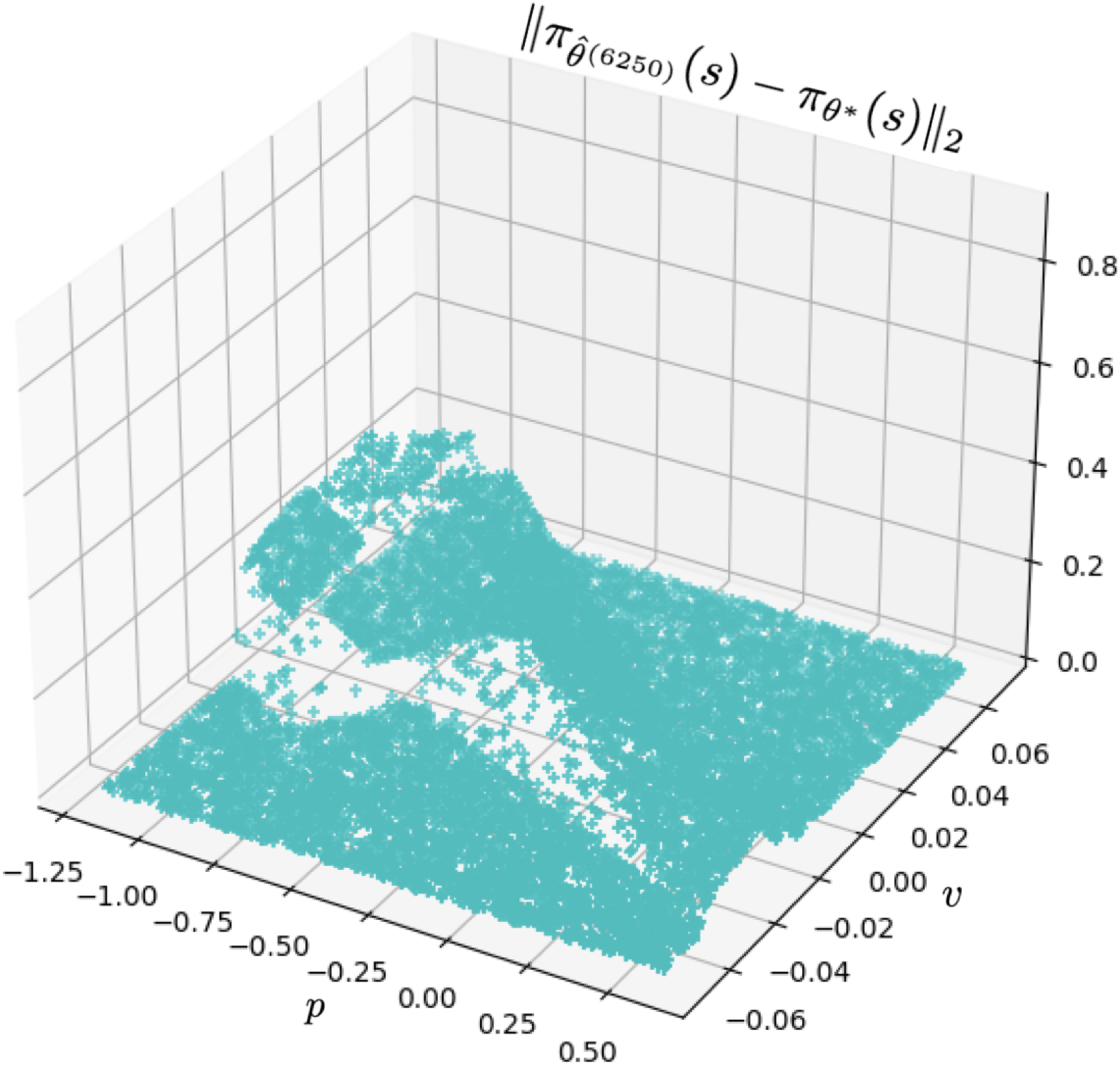}
    \includegraphics[width=0.326\textwidth]{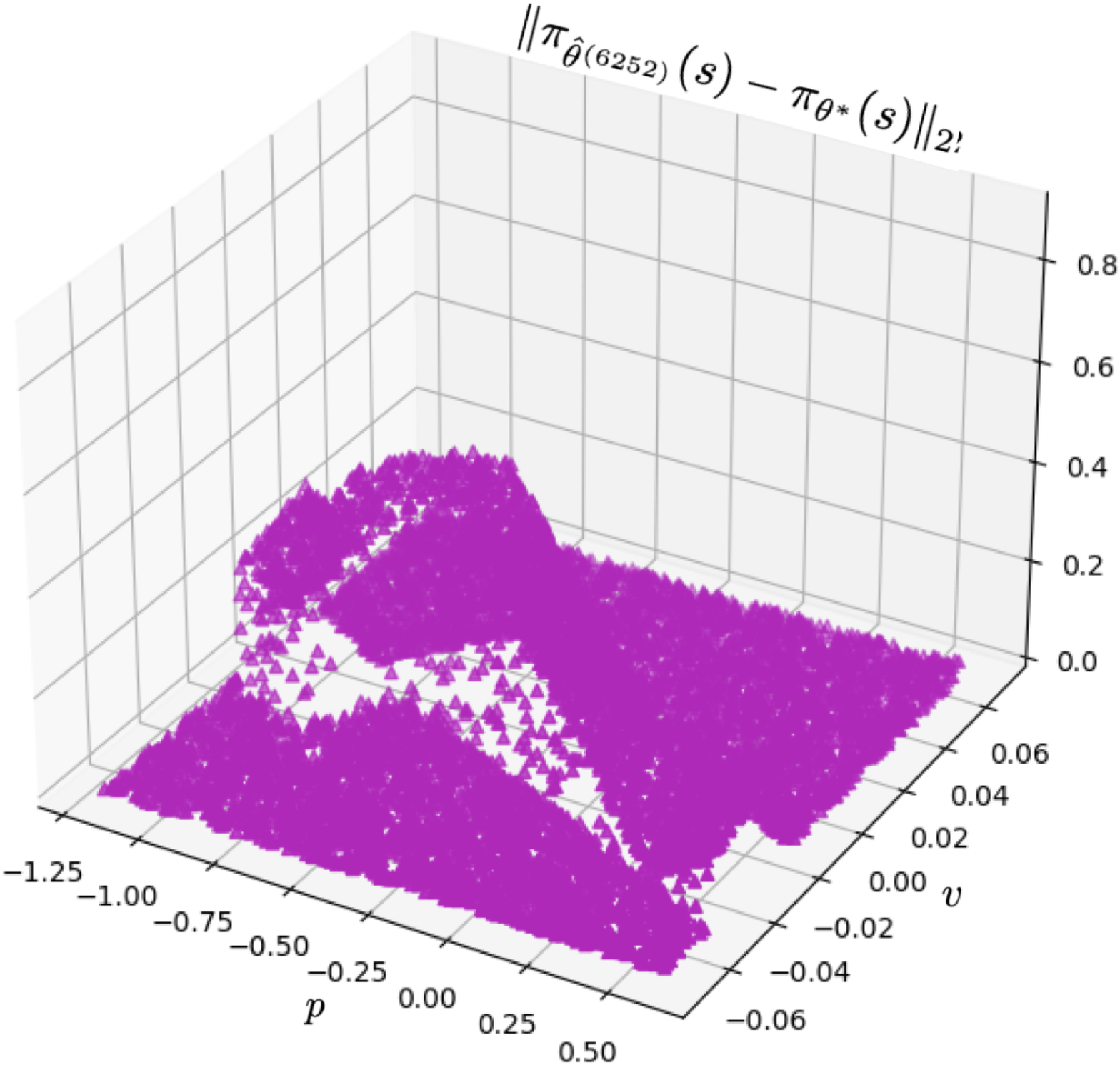}
    \includegraphics[width=0.326\textwidth]{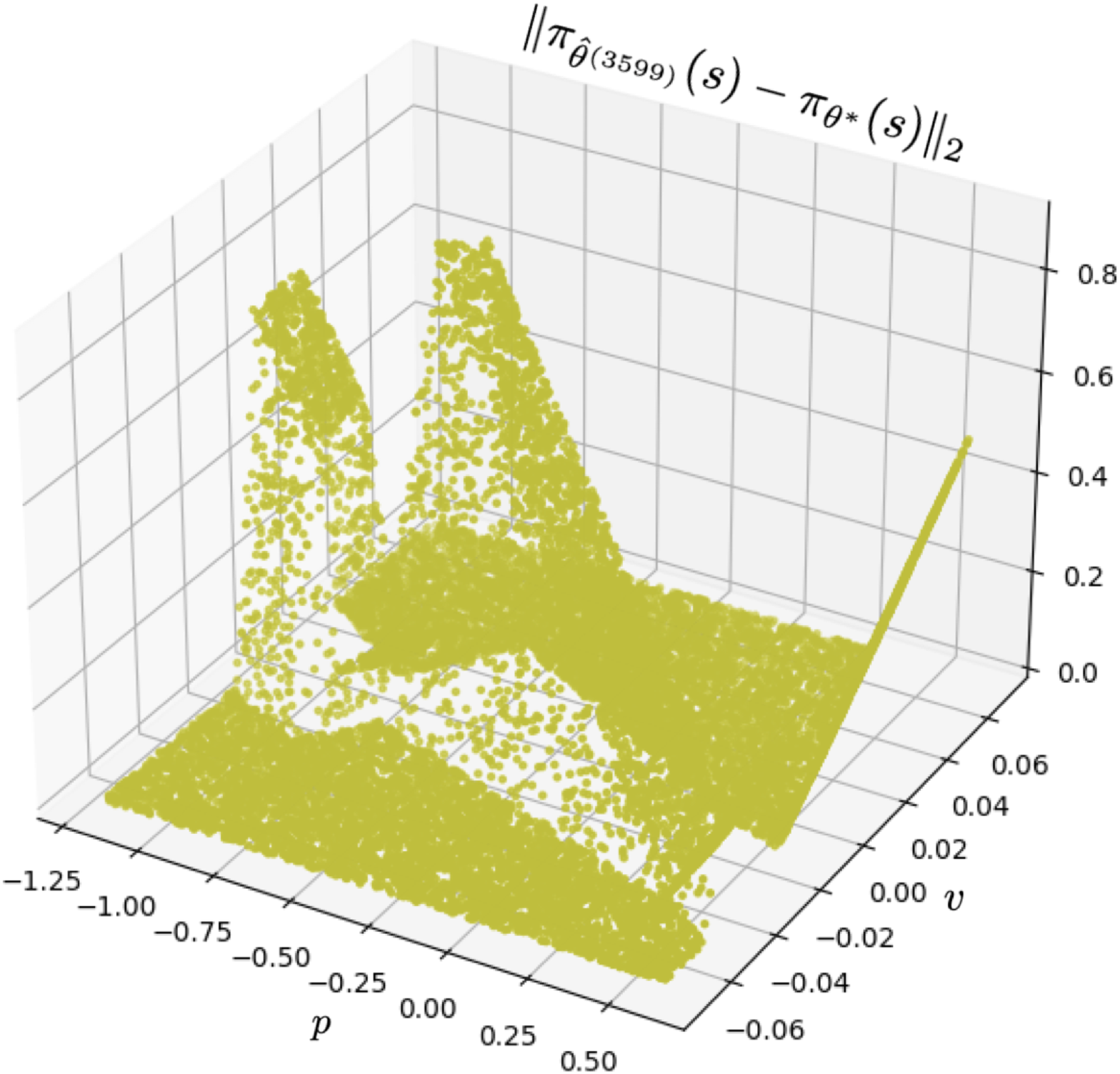}
    \caption{Error $\| {\pi_{\hat{\theta}}}({s}) - \pi_{{{\theta}}^{*}}({s}) \|_2$ , over uniform samples across the state space of ${\pi_{\hat{\theta}}} \in  \pi_\text{Str}$ generated by Algorithm \ref{alg:online}. The global candidate policies $\{\pi_{\hat{\theta}^{6250}}, \pi_{\hat{\theta}^{6252}}\}$ are always well below the user threshold $\psi=0.55$. 
    }
    \label{fig:err_exp2}
\end{figure*}

Figure \ref{fig:traj_err_exp2} shows the results of one of these runs, which resulted in $\pi_\text{Str} = \{\pi_{\hat{\theta}^{6250}}, \pi_{\hat{\theta}^{6252}}, \pi_{\hat{\theta}^{3599}}\}$. In particular, Figure \ref{fig:traj_err_exp2} shows the error $\| {\pi_{\hat{\theta}^{(i)}}}({s}^{(k)}) - \pi_{{{\theta}}^{*}}({s}^{(k)}) \|_2$ across the trajectory of collected observations for all shortlisted policies. As can be observed from the figure, all candidates maintain an error that is below the threshold of $\psi=0.55$. In Figure \ref{fig:err_exp2}, we randomly sample across the entire state space and plot the error for the shortlisted candidate policies $\pi_\text{Str}$. As promised by Corollary~\ref{th:online}, at least one of the identified policies has a maximum error that is less than $\psi=0.04$. More specifically the maximum error for ${\pi_{\hat{\theta}^{(6250)}}} = 0.463$. 

Figures \ref{fig:traj_err_exp2} and \ref{fig:err_exp2} highlight our assumption on the attacker's capabilities in Section II.C and, in particular, that we place no requirements on the quantity and coverage of the observations $\mathcal{D}$. As seen in Figure~\ref{fig:traj_err_exp2}, policy $\pi_{\hat{\theta}^{3599}}$ was shortlisted because it met the attacker threshold $\psi$ across this trajectory. Nevertheless, the policy $\pi_{\hat{\theta}^{3599}}$ does not, in general, satisfy the threshold $\psi$ as shown in Figure~\ref{fig:err_exp2}.

\section{Conclusion}
In this paper, we introduced the model extraction attacks on reinforcement-learning-based neural network controllers. The attacker's objective is to estimate an unknown neural network controller to predict its future control outputs. We proposed a two-phase algorithm. In the first phase, we produce a set of candidate estimates of the unknown controller using the side-channel information about the optimization problem originally used to design that controller. We then use a limited set of the observed state/control action pairs in the online phase to shortlist the estimates from the offline phase. We provided a theoretical analysis that establishes the correctness of our algorithms. Our numerical experiments on two different sets of tasks show that only a limited amount of observations is needed to produce estimates that are close enough to the unknown controller.
\vspace{5mm}

\bibliographystyle{IEEEtran} 
\bibliography{main_bib}
\appendix
%\vspace{-3.5mm}
\subsection{Proof of Proposition~\ref{prop:1}}
\begin{proof} 
In order to bound the error $ \max_{\theta \in \Theta} \|g_j(\theta) - \widehat{g}_j(\theta)\|_2 $, we split it into the two parts below:
\begin{align}
 \vspace{1mm}
    \max_{\theta \in \Theta} \|g_{j}(\theta) - \widehat{g}_{j}(\theta)\|_2
    &
    \le \max_{\theta \in \Theta} \|g_{j}(\theta )- \tilde{g}_{j}(\theta) \|_2 \nonumber \\ 
    & \qquad + \max_{\theta \in \Theta} \|\widehat{g}_{j}(\theta)  - \tilde{g}_{j}(\theta)\|_2
     \vspace{1mm}
    \label{eq:bound_vector_component}
\end{align}

To bound the first term on the RHS of \eqref{eq:bound_vector_component}, we follow the same argument used in proving the error bound for sampled derivatives of real-valued functions in \cite{yano2012math} as follows:

\begin{align}
\displaystyle\ \max_{\theta \in \Theta}&
\| {g}_{j}({\theta}) - \widetilde{g}_{j}({\theta})\|_2 \notag 
\\ 
&
\stackrel{(a)}{=} \displaystyle \max_{\theta \in \Theta} \|{g}_{j}({\theta}) - \frac{J({{\theta}}+c_j) - J({{\theta}})}{c} \notag
\\ 
&  \qquad \qquad \qquad \qquad + \frac{ m(\tilde{\theta}) - m(\tilde{\theta} + c_j)}{c}\|_2  \notag
\\ 
& \stackrel{(b)}{\leq} \max_{\theta \in \Theta} \|{g}_{j}({\theta}) - \frac{J({{\theta}}+c_j) - J({{\theta}})}{c} \| \notag 
\\ 
& \qquad \qquad \qquad \qquad + \displaystyle \max_{\theta \in \Theta} \| \frac{m(\tilde{\theta}) - m(\tilde{\theta} + c_j)}{c}\|  \notag
\\ 
& \stackrel{(c)}{\leq} \max_{\theta \in \Theta} \|{g}_{j}({\theta}) - \frac{J({{\theta}}+c_j) - J({{\theta}})}{c} \| + \frac{2\overline{m}}{c}  \notag
\\ 
& \stackrel{(d)}{=} \max_{\theta \in \Theta} \|{g}_{j}({\theta}) - \frac{1}{c}( g_{j}({\theta})c + \frac{1}{2}g_{j}'({\Lambda})c^{2}\|_2 + \frac{2\overline{m}}{c}  \notag
\\ 
& = \|\frac{1}{2}g_{j}'({\Lambda})c\|_2 + \frac{2\overline{m}}{c}  \notag
\\ 
& {\leq} \frac{1}{2}  \max_{\|{{\Lambda}\| \in [ \|{\theta}\|, \|{\theta} + c_{j}\|]}} \|{g_{j}}'({\Lambda})\|_2 c + \frac{2\overline{m}}{c} \notag
\\ 
& \stackrel{(e)}{\le} c {L}_{g_{j}} + \frac{2\overline{m}}{c}  
\label{eq:component_1}
\end{align}

where $g_{j}'$ is the derivative of $g_{j}$ and ${\Lambda} \in \mathbb{R}^l$ where $\|{\Lambda}\| \in [  \|{\theta}\|,  \|{\theta + c_{j}}\| ]$. Equation $(a)$ follows from the definition of $\widetilde{g}_{j}({\theta})$ in equation \eqref{eq:gtilda}. Inequality $(b)$ uses triangular inequality, and  equation $(c)$ uses the bound on $m(\theta)$ as  defined in~\eqref{eq:bound_m}. Equation $(d)$ follows from assumptions 2 and 3 in the proposition, which ensures that the function $J$ is twice differentiable and hence once can apply the Taylor series expansion and the Mean-Value Theorem, which ensures the existence of a $\Lambda$ such that $\|{\Lambda}\| \in [  \|{\theta}\|,  \|{\theta + c_{j}}\| ]$ for which equality holds. The inequality in $(e)$ follows from assumptions 1-3 in the proposition, which ensure that $g$ is Lipschitz continuous~\cite[Lemma~4.1]{shen2019hessian} with 
$L_{g} \le \frac{H^2 G^2 R^2 + L^2 R^2}{1-\gamma^4}$. 

Next, we bound the second term on the RHS of~\eqref{eq:bound_vector_component}:% as follows:
\begin{align}
    \max_{\theta \in \Theta} & \|\widehat{g}_j(\theta)  - \tilde{g}_j(\theta)\|_2 
     \stackrel{(f)}{\le} \nonumber 
    \\
    & \max_{\theta \in \{\tilde{\theta}^{(1)}, \ldots, \tilde{\theta}^{(M)} \} } \|\widehat{g}_j(\theta)  - \tilde{g}_j(\theta)\|_2
    \nonumber 
    \\
    & \qquad \qquad + \max_{\theta \in \Theta \setminus \{\tilde{\theta}^{(1)}, \ldots, \tilde{\theta}^{(M)} \} } \|\widehat{g}_j(\theta)  - \tilde{g}_j(\theta)\|_2 
    \nonumber 
    \\
    &
    \stackrel{(g)}{=} \zeta_j + \max_{\theta \in \Theta \setminus \{\tilde{\theta}^{(1)}, \ldots, \tilde{\theta}^{(M)} \} } \|\widehat{g}_j(\theta)  - \tilde{g}_j(\theta)\|_2 
    \nonumber 
    \\
    &
    \stackrel{(h)}{\le} \zeta_j + \eta L_{\xi_j}
     \label{eq:lipschtiz_xi}
\end{align}
where $(f)$ follows from splitting the error into two parts, one at the training samples $\tilde{\theta}^{(1)}, \ldots, \tilde{\theta}^{(M)}$ and another one that represents the error between the training samples. Equation $(g)$ follows from the definition of $\zeta_j$ as the maximum sample error. Assuming the function $\xi_j(\theta) = \widehat{g}_j(\theta)  - \tilde{g}_j(\theta)$ is Lipschitz continuous with a Lipschitz constant $L_{\xi_j}$, the maximum error between any two samples is bounded by $\eta L_{\xi_j}$ where $\eta$ is the grid size used to sample $\tilde{\theta}^{(1)}, \ldots, \tilde{\theta}^{(M)}$ and hence inequality $(h)$ holds.

What is remaining is to show that the function $\xi_j$ is Lipschitz continuous. Nevertheless, it follows from the definition of $\xi_j$ that it is the summation of two functions $\widehat{g}_j$ and $\tilde{g}_j$. The first function $\widehat{g}_j$ is the linear combination of kernel functions that are Lipschitz continuous and hence $\widehat{g}_j$ is Lipschitz continuous with a constant $L_{\widehat{g}_j}$. Similarly, $\tilde{g}_j$ is the sample approximation of the continuous function $g$ (thanks to the assumptions 1-3) and hence Lipschitz continuous with a constant $L_{\tilde{g}_j}$.

Thus, the Lipschitz constant $L_{\xi_j}$ of $\xi_j$ is equal to $(L_{\widehat{g}_j} + L_{\tilde{g}_{j}})$. 
Substituting in equation
\eqref{eq:lipschtiz_xi}, we conclude that:
\begin{align}
    \max_{\theta \in \Theta} \|\widehat{g}_j(\theta)  - \tilde{g}_j(\theta)\|_2 \le \zeta_j + \eta (L_{\widehat{g}_j} + L_{\tilde{g}_{j}}) \label{eq:component_2}
\end{align}

The overall component-wise bound is then found by substituting equations \eqref{eq:component_1} and \eqref{eq:component_2}  in equation~\eqref{eq:bound_vector_component} as:% below, where $j \in \{1, \ldots, l\}$: 
\begin{align}
\max_{\theta \in \Theta} \|g_{j}(\theta) - \widehat{g}_{j}(\theta)\|_2 \le \underbrace{c {L_{g_{j}}} + \frac{2\overline{m}}{c} + \zeta_{j} + \eta ({L_{\widehat{g}_{j}}} + {L_{\tilde{g}_{j}}})}_{e_{j}}
\label{eq:component_e}
\end{align}

\end{proof}

\subsection{Proof of Proposition~\ref{prop:2}}

\begin{proof}

We start by defining the sets below:
\begin{align}
    % & S_{{\theta}^*}=\{{\theta} | \hspace{0.1cm} g({\theta}) = 0\} \label{eq:Set_thetastar}
    & S_{{\theta}^*}=\{{\theta} \; | \; g_{j}({\theta}) = 0, \; j \in \{1, \ldots, l\}\} \label{eq:Set_thetastar}
    % \\ 
    % &  
    % S_{{\overline{\theta}}}=\{{\theta} | \hspace{0.1cm} -{\bar{e}} \leq \widehat{g}({\theta}) \leq \bar{e}\} \label{eq:Set_thetabar}
    \\
    % &
    % S_{{\overline{\theta}}}=\{{\theta} \; | \; - \overline{m} \leq \overline{g}_{j}({\theta}) \leq \overline{m}||, \; j \in \{1, \ldots, l\}\} \label{eq:aaa}
    %  \\
     &
    S_{{\widehat{\theta}}}=\{{\theta} \; | \; -{{e}_{j}} \leq \widehat{g}_{j}({\theta}) \leq {e}_{j}, \; j \in \{1, \ldots, l\}\} \label{eq:Set_thetabar}
    \\ 
    & 
    S_{B_{{\widehat{\theta}}}}= \mathbb{B}_b({\hat{\theta}}^{(1)}) \cup \mathbb{B}_b({\hat{\theta}}^{(2)}) \cup ... \cup \mathbb{B}_b({\hat{\theta}}^{(r)}) \label{eq:Union}
\end{align}

Indeed, it is direct to show that equation~\eqref{eq:prop2} is equivalent to showing that $\theta^* \in S_{B_{{\widehat{\theta}}}}$. To show the correctness of this statement, we aim to prove the following sequence of set inclusion:
\begin{align}
\theta^* \stackrel{(a)}{\in} S_{{\theta}^*} %\bb{\stackrel{(b)}{\subseteq} S_{{{\theta'^*}}}} 
\stackrel{(b)}{\subseteq} S_{{\widehat{\theta}}} \stackrel{(c)}{\subseteq} S_{B_{{\widehat{\theta}}}}
\end{align}
The set membership $(a)$ follows from the definition of $S_{{\theta}^*}$ along with the fact that $\theta^*$ is a critical point of $J$ (recall equation~\eqref{eq:theta_star}) and hence $g(\theta^*) = \frac{d}{d \theta} J(\theta) |_{\theta = \theta^*}$ is equal to zero. %The set inclusion $(b)$ follows from the definition of $\overline{g}$ when $g(\theta)=0$.% as defined in ~\eqref{eq:actual_g} is equivalent to saying that the function we are observing $g'(\theta)$, as defined in \eqref{eq:obs_g}, is within $\pm$ m.
The set inclusion $(b)$ can be shown as follows. First, we note that the following holds for any $\theta \in S_{{\theta}^*}$:
\begin{align}
\|{g}_{j}({\theta}) - \widehat{g}_{j}({\theta})\|_2 \stackrel{(e)}{=} \| \widehat{g}_{j}({x})\|_2 \stackrel{(f)}{=} | \widehat{g}_{j}({x}) | \stackrel{(g)}{\le} e_j
\label{eq:set_membership_2}
\end{align}
where $(e)$ follows from the definition of $S_{{\theta}^*}$ which ensures that ${g}_{j}({\theta}) = 0$, $(f)$ follows from the fact that $\widehat{g}_{j}({x})$ is a scalar and hence $\| \widehat{g}_{j}({x})\|_2 = \sqrt{(\| \widehat{g}_{j}({x})\|_2)^2} = |\widehat{g}_{j}({x})|$, and $(g)$ follows from~\eqref{eq:component_e}. Hence we conclude that each $\theta \in S_{{\theta}^*}$ satisfies the conditions of $S_{{\widehat{\theta}}}$ which proves $(b)$ above. Finally, the set membership in $(c)$ follows from equation \eqref{eq:smt1}, which concludes the proof.

\end{proof}
\vspace{1mm}
\subsection{Proof of Theorem~\ref{th:offline}} 
\begin{proof}
It follows from the assumptions in Proposition \ref{prop:1} that $\pi$ is a Lipschitz continuous function with a Lipschitz constant $G$ and hence the following holds for any $s \in S$:
\begin{align}
    \|  \pi_{{\theta}^{*}}(s) -  {\pi}_{{\hat{\theta}}^{(i^*)}}(s)\|_2
    & \stackrel{(a)}{\le} G \|{\theta}^{*} - {\hat{\theta}}^{(i^{*})} \|_2 \nonumber
    \\
    &
    \stackrel{(b)}{\le}  
    G b \nonumber
    \\ 
    & 
    = \psi \nonumber 
\end{align}
where $(a)$ is implied by assumption (2) in Proposition \ref{prop:1} and $(b)$ follows from Proposition \ref{prop:2}.
\end{proof}

\end{document}